\pdfoutput=1

\documentclass[11pt]{article}

\usepackage[preprint]{acl}

\usepackage{times}
\usepackage{latexsym}

\usepackage[T1]{fontenc}

\usepackage[utf8]{inputenc}

\usepackage{microtype}

\usepackage{inconsolata}

\usepackage{graphicx}

\title{Spoken Conversational Agents with Large Language Models}

\author{Chao-Han Huck Yang$^{1}$\quad Andreas Stolcke$^{2}$
\quad Larry P Heck$^{3}$ \\
 NVIDIA Research$^{1}$ \quad Uniphore$^{2}$\quad Georgia Institute of Technology$^{2}$\\
{\tt\small hucky@nvidia.com \quad andreas.stolcke@uniphore.com \quad larryheck@gatech.edu }\\
}

\begin{document}
\maketitle
\section{Introduction}

Recent advancements in large language models (LLMs) with \textit{voice interfaces} have garnered significant attention from both the research community and broader society. Closed-source models, such as GPT-4o and Gemini-1.5-pro, have demonstrated superior performance across classical speech tasks, including (i) speech recognition, (ii) translation, and (iii) spoken language understanding, significantly surpassing previous open-source benchmarks. Despite these strides, there remains a lack of comprehensive studies on the design and mechanisms underpinning the integration of speech modalities into LLMs for true multi-modal understanding. Additionally, the interaction between speech models and LLMs, particularly in the context of layered, self-play cognitive agents~\cite{shah-etal-2018-bootstrapping, shah2018buildingconversationalagentovernight}, has only recently begun to be explored.

This tutorial aims to provide a thorough review of the historical trajectory of probabilistic language modeling~\cite{jurafskymartin} for speech processing, offering insights that motivate the development of multi-agents system in conversational models. We will delve into advanced topics such as cross-modal adaptation, introducing on the theoretical foundations~\cite{yang2021voice2series} required to align non-text modalities with textual representations. These concepts will be discussed alongside open-source, reproducible benchmarks~\cite{chen2023hyporadise}, providing a practical grounding for participants.

\begin{figure}[htbp]
\centering
\includegraphics[width=0.42\textwidth]{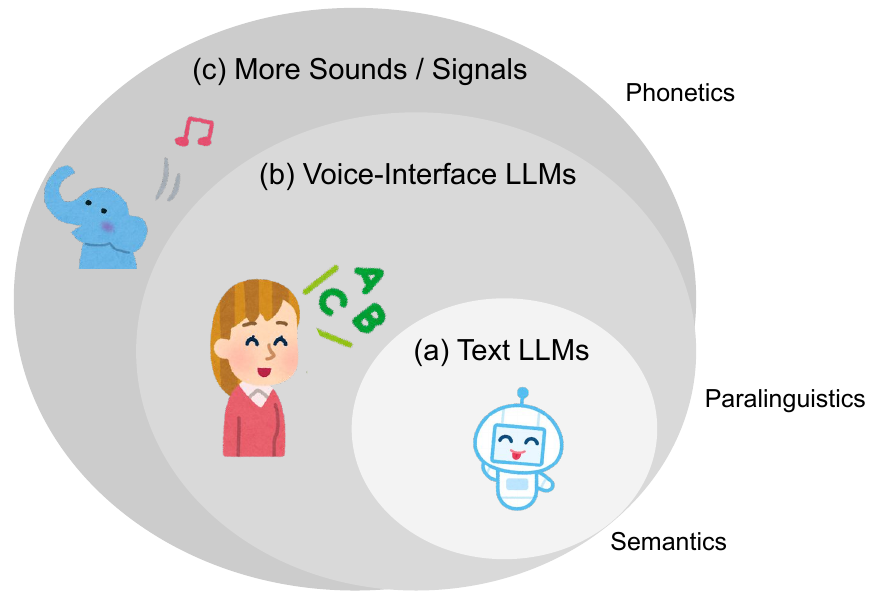} \\
\caption{Examples of spoken conversational agent with different LLMs to understand different linguistic information from (a) semantics, (b) paralinguistics, and to (c) more phonetic signals.}
\label{fig:exp}
\end{figure}

In addition, we will explore more recent trends toward end-to-end multi-modal speech-language models, emphasizing designs that utilize generative autoregressive approaches with joint speech-text tokenization. By examining both cascaded and end-to-end perspectives, this tutorial will equip participants with a comprehensive understanding of current strategies and open challenges in speech-augmented LLMs as shown in Figure~\ref{fig:exp}.





\section{Tutorial Outline}
This three-hour tutorial will focus on the rapidly evolving field of speech-language modeling in the era of voice-interfacing LLMs and agent systems. Each core theme will be covered in a $35$-minute segment, followed by a $10$-minute Q\&A and a $10$-minute break. For each section, we will provide an overview of the relevant topics, followed by an in-depth exploration of key studies that shape the current landscape. The tutorial will conclude with a discussion of the open challenges and emerging research opportunities in this exciting and transformative area of joint speech-language modeling.

\subsection{Language Modeling for Speech Processing Background, History, and Beyond}
\begin{itemize}
\item \textbf{Probabilistic LMs for Speech Signals:} We will begin by discussing the early foundational work in Bayesian and n-gram-based language models, which were instrumental in building probabilistic frameworks for speech recognition tasks. These models laid the groundwork for current systems by modeling word sequences and handling uncertainties in speech inputs. 

\item \textbf{Contextual End-to-End Speech Models:} Next, we will explore the rise of end-to-end speech models that integrate contextual information directly into the modeling process. These models capture not only linguistic content but also paralinguistic features such as prosody, emotion, and speaker characteristics, allowing for more robust and nuanced speech understanding and generation. 

\item \textbf{Post-ASR LLM Correction and Fairness:} Finally, we will examine the role of LLMs in post-ASR error correction, where language models are used to refine and correct the outputs of traditional ASR systems. Special attention will be given to the fairness and bias issues that arise in speech processing, particularly with regard to speaker variability (e.g., accent, dialect, and sociolect). We will discuss how recent advances are addressing these challenges, and the ethical implications of ensuring equitable performance across diverse speech populations.

\end{itemize}

\subsection{Large Language Models for Audio, Speech, and Conversational Signals}

\begin{itemize}
    \item \textbf{Theoretical Basics of LLMs Adaptation}: We will review some theoretical foundations of LLM adaptation and how these frameworks connect to LLMs with speech input. Topics include population risk measurement~\cite{yang2021voice2series} and model transferability estimation~\cite{chen2023estimate} from speech models to LLM adaptation, motivating different design pipelines~\cite{radhakrishnan2023whispering,hu-etal-2024-gentranslate, chen2024s, yang2023generative} of spoken agents.
    \item \textbf{Speech-Text Pre-training / Post-alignment}: Building on this, we will examine joint text-speech pre-training~\cite{chiu2022self, barrault2023seamless, chen2022maestro} methods, which have pushed the boundaries of multi-modal understanding by combining speech and text learning objectives. The application of LLMs for voice quality estimation will also be discussed, showing how these models can assess and adapt to different speaker characteristics in real-time.
    \item  \textbf{Multi-task Evaluation for Voice-LLMs:} Lastly, we will cover the latest advances in joint generative translation models, which integrate both speech and text modalities for seamless, high-quality translation across languages. This section will provide a comprehensive look at how state-of-the-art voice-interfaced LLMs~\cite{reid2024gemini,chu2023qwen, radford2023robust} are the processing and understanding of speech and conversational signals.
\end{itemize}

\subsection{Spoken Dialogue Systems}

\begin{itemize}
    \item {\bf Historical Foundations} The last decade saw significant progress in the creation and deployment of large-scale voice-powered AI virtual assistant technology, starting with Siri (SRI/Apple), and then followed by Cortana (Microsoft), Google Assistant, Alexa (Amazon), and Viv/Bixby (Samsung). This section of the tutorial will cover the technological innovations underlying this era of AI virtual assistants including intent detection and slot filling with RNNs \cite{mesnil2014using}, dialogue management with reinforcement learning \cite{shah2016}, leveraging web-scale query-clicks \cite{hakkani2011exploiting, tur2011sentence}, exploiting the semantic web \cite{heck2012exploiting}, and incorporating knowledge graphs in spoken language understanding \cite{huang2015leveraging}
    \item {\bf Current Trends}
   The current work in AI virtual assistants builds upon the voice-only systems of the last decade by leveraging LLMs to significantly improve the coverage and robustness of the spoken language understanding and dialogue state tracking components, in addition to substantial advancements in spoken language generation. This tutorial section provides an overview of existing LLMs and methodologies for adapting them to downstream tasks \cite{Ni2021RecentAI, qin-etal-2023-end, Yi2024ASO}. It highlights recent advancements in multi-turn dialogue systems, encompassing both LLM-based open-domain dialogue (ODD) and task-oriented dialogue (TOD) systems, as well as relevant datasets and evaluation metrics. Additionally, it addresses emerging research challenges associated with the development of LLMs and the growing demands on multi-turn dialogue systems.
   \item {\bf Future Directions}
   Future work will build on the recent progress in LLMs and unify task-oriented and open-domain dialogue systems. This section of the tutorial will cover this technology shift and highlight some examples of promising future directions. Some examples include building new foundation LLMs pretrained with conversational data \cite{jawale2024human}, e2e training of dialogues \cite{liu2017end, liu2018dialogue}, and dialogue self-play \cite{shah-etal-2018-bootstrapping, shah2018buildingconversationalagentovernight}.
  Another significant future direction is situated dialogue systems: grounding LLM-based dialogue systems with content (web pages \cite{heck2013multi}, lists \cite{bapna2017towards}, forms \cite{heck2024mforms}, tables \cite{sundar-heck-2023-ctbls, sundar2024gtbls, sundar2024itbls}, papers with figures, equations, tables \cite{sundar2024cpapers}, and retrieved documents \cite{reichman2024retrieval}, vision \cite{hakkani2014eye}, knowledge \cite{reichman2023outside}, emotion \cite{reichman2024reading}, and expression including facial/lip movement, facial/body expressions, and gestures \cite{punjwani2024largebodylanguagemodels, punjwani2024alloavalargescalemultimodalconversational}.
\end{itemize}

\section{Tutorial Presenters}

\paragraph{Huck Yang} is a Senior Research Scientist at NVIDIA Research. He received his PhD degree from Georgia Institute of Technology. His research focuses on speech-language modeling, robust speech recognition, and multimodal post-editing models. He gave a tutorial at ASRU and Intetspeech 2023  on ``Parameter-Efficient Language Modeling for Speech Processing,'' and a series of tutorials at ICASSP on ``Prompt Learning for Speech-Language Models'' from 2022 to 2024. He has served as area chairs in Interspeech, SLT, and ICASSP and worked full time at Amazon AGI. 

\paragraph{Andreas Stolcke} is a Distinguished AI Scientist and VP at Uniphore. He holds a PhD in Computer Science from UC Berkeley and has previously held positions at Amazon, SRI International, and Microsoft. His research spans language modeling, speech recognition, speaker identification, and speech translation. He is an IEEE Fellow and a Fellow of the International Speech Communication Association (ISCA), with long-term research interests in language modeling for speech processing, paralinguistics, and conversational AI.
\paragraph{Larry Heck} is a Professor with a joint appointment in ECE and Interactive Computing, Chief Scientist of Tech AI, and Executive Director of the Machine Learning Center at the Georgia Institute of Technology. He holds the Rhesa S. Farmer Distinguished Chair of Advanced Computing Concepts and is a Georgia Research Alliance Eminent Scholar. He received a PhD in Electrical Engineering from Georgia Institute of Technology, is an IEEE Fellow, and has previously held positions at SRI International, Nuance, Yahoo!, Microsoft, Google, Viv Labs, and Samsung. His career includes research on acoustics, active noise and vibration control, speaker recognition, web search, AI virtual assistants, NLP, and conversational AI.

\section{Reading List and Prerequisite}
This tutorial is designed for researchers and practitioners working at the intersection of conversational agents, and spoken information interactions. We assume attendees will have a foundational understanding of NLP and speech processing. While experience with system deployment is beneficial, we will provide a carefully paced introduction to core materials to accommodate a broader audience. Below are a few papers that offer important insights and foundations for this tutorial:

\begin{itemize}
\item Dialogue act modeling for automatic tagging and recognition of conversational speech~\cite{stolcke2000dialogue}
    \item Is spoken language all-or-nothing? Implications for future speech-based human-machine interaction~\cite{moore2017spoken}
    \item Dialogue Learning with Human Teaching and Feedback in End-to-End Trainable Task-Oriented Dialogue Systems~\cite{liu2018dialogue}
    \item Voice2series: Reprogramming acoustic models for time series classification~\cite{yang2021voice2series}
    \item Hyporadise: An open baseline for generative speech recognition with large language models~\cite{chen2023hyporadise}
    \item SpeechGPT: Empowering Large Language Models with Intrinsic Cross-Modal Conversational Abilities~\cite{zhang2023speechgpt}
\end{itemize}

\paragraph{Breadth} While we will reference dozens of relevant papers throughout the tutorial, we plan to take a closer look at 7-8 key research papers in detail. Of these, only 1-2 will be directly authored by the presenters, ensuring a broad and balanced exploration of the field.








\section{Diversity Considerations}

As we advance toward more conversational agents with LLMs, addressing diversity from spoken information is not just a matter of fairness, but of achieving the robustness and versatility necessary for real-world applications. Voice-interface based LLMs have made significant progress, but their limitations in handling diverse linguistic variations—accents, dialects, sociolects -- present both ongoing challenges and opportunities for improvement.

\begin{itemize}
    \item \textbf{Accents and Dialects: A Multidimensional Challenge}

    Speech processing models have traditionally struggled with accent and dialect diversity. While LLMs have shown impressive performance, they still reflect the biases of the datasets on which they are trained. These models often prioritize standard accents or high-resource languages, leading to suboptimal performance for speakers with less-represented accents. Moving forward, we need to systematically address this by designing models that generalize across diverse phonetic and prosodic patterns. Studies such as in-context learning based adaptation and retrieval-augmented clustering could play a crucial role in making LLMs more resilient to linguistic variation, allowing the models to dynamically adapt to previously unseen accents with minimal data.

    \item \textbf{Redefining Diversity-Oriented Evaluation Metrics for Spoken Conversational Agents}

    Standard evaluation metrics for LLMs in speech processing—such as word error rate (WER) for ASR—are often insufficient for capturing the true performance disparities across diverse linguistic groups. We must expand our evaluation frameworks to include fairness-driven metrics that assess model performance across different demographics, accents, and languages. Additionally, error analysis should go beyond quantitative measures and include qualitative insights into which user groups are more affected by model biases. Such nuanced evaluation will help guide the development of more equitable models.



    \item \textbf{Sociolects and Variability Across Social Dimensions for Conversational Agents}

   The diversity within a language itself—manifested through sociolects—poses an additional layer of complexity. Speech patterns vary based on age, gender, socio-economic background, and even profession. Traditional speech models tend to overfit to more homogenous data distributions, often reflecting the sociolects of dominant groups in the training datasets. Addressing this requires more than just adding diverse data; it demands sophisticated mechanisms like style transfer and speaker adaptation that allow models to process a wide spectrum of linguistic and paralinguistic cues. In particular, attention to prosody, intonation, and speaker emotion is key to better understanding socially or emotionally charged speech.
\end{itemize}

\section{Ethics Statement}

As we advance the integration of LLMs with spoken interactions, it is essential to proactively address the ethical implications associated with these developments. In this tutorial, we aim to raise awareness of these issues while equipping participants with strategies for responsible innovation. To reach a broad and diverse audience, we will promote the tutorial across various platforms. Our presenters include both \textit{early-career }and \textit{senior} researchers from both industry and academia. On speech privacy and security consideration, with the growing ubiquity of voice-interfaced systems in everyday life, both in public and private domains, protecting users' speech data from potential misuse is of critical importance. Voice data can contain sensitive personal information, and ensuring its secure handling is non-negotiable. 

Our team brings substantial expertise in this area. For instance, Huck has served on the IEEE data collection committee, reviewing voice and signal data benchmarks with a focus on ethical data use. Andreas and Larry, both IEEE Fellows, have extensive experience in language modeling for speech processing, including speaker identification, which presents unique privacy challenges. Andreas gave a plenary talk on ``\textit{Speech-based and Multimodal Approaches for Human versus Computer Addressee Detection}'' at EMNLP 2016.

\bibliography{custom}




\end{document}